\pdfoutput=1
\documentclass[11pt]{article}
\usepackage{graphicx}

\usepackage{acl}
\usepackage{amsmath}
\usepackage{mathtools}
\usepackage{amssymb}
\usepackage{times}
\usepackage{latexsym}
\usepackage{booktabs}
\usepackage{multirow}
\usepackage{adjustbox}
\usepackage{siunitx}

\usepackage{array}
\newcolumntype{P}[1]{>{\centering\arraybackslash}p{#1}}

\usepackage{algorithm}
\usepackage{algpseudocode}

\usepackage[T1]{fontenc}
\usepackage[utf8]{inputenc}

\usepackage{microtype}

\title{Nowruz at SemEval-2022 Task 7: Tackling Cloze Tests with Transformers and Ordinal Regression}

\author{
  Mohammadmahdi Nouriborji \\
  NODET, Iran\footnotemark\\
  \texttt{mohammadmahdinoori70@gmail.com} \\\AND
  Omid Rohanian \\
  Department of Engineering Science \\
  University of Oxford \\
  Oxford, UK \\
  \texttt{omid.rohanian@eng.ox.ac.uk} \\\And
  David Clifton \\
  Department of Engineering Science \\
  University of Oxford \\
  Oxford, UK \\
  \texttt{david.clifton@eng.ox.ac.uk} \\
}

\begin{document}
\maketitle
\footnotetext{National Organization For Development of Exceptional Talents}
\begin{abstract}
This paper outlines the system using which team Nowruz participated in SemEval 2022 Task 7 “Identifying Plausible Clarifications of Implicit and Underspecified Phrases” for both subtasks A and B \citep{roth-etal-2022-identifying}. Using a pre-trained transformer as a backbone, the model targeted the task of multi-task classification and ranking in the context of finding the best fillers for a cloze task related to instructional texts on the website Wikihow. 

The system employed a combination of two ordinal regression components to tackle this task in a multi-task learning scenario. According to the official leaderboard of the shared task, this system was ranked 5th in the ranking and 7th in the classification subtasks out of 21 participating teams. With additional experiments, the models have since been further optimised. The code used in the experiments is going to be freely available at \url{https://github.com/mohammadmahdinoori/Nowruz-at-SemEval-2022-Task-7}.  
\end{abstract}

\section{Introduction}
Oxford dictionary defines cloze test as ``a test of readability or comprehension in which a person is required to supply words which have been deliberately omitted from a passage'' \citep{oed:cloze}. In the context of NLP, a cloze format task is one in which the context is one or more sentences with  masked spans and the model is expected to predict a suitable filler for each span. Cloze-format datasets have become popular in NLP recently as they are relatively easy to create automatically and provide high quality resources for model training \citep{rogers2021qa}. 

SemEval 2022 task 7 is framed as a cloze task in which the goal is to rank or classify fillers within a given context based on their suitability. The texts are taken from actual articles on an instructional website and the masked spans are placed at locations of edits made by users. We participated in this shared task in both ranking and classification parts and developed a transformer-based model that utilises both classification and regression components at the top layer. The code and the data used in these experiments are publicly available.

% The rest of this paper is structured as follows: In Section \ref{related-work} we will look at some relevant literature. Section \ref{system-desc} provides an overall description of the system and detailed explanation of each component. Dataset and the experimental setup are described in Section \ref{experiments}. In Section \ref{results} the results are shared and discussed. Section \ref{error-analysis} is about error analysis and finally Section \ref{Conclusion} concludes the paper and outlines some possible future research directions. 

\section{Related Work}
\label{related-work}

Cloze tasks have been a subject of interest in Natural Language Understanding (NLU) in recent years, especially within the context of reading comprehension, story understanding, and summarisation \citep{deutsch-roth-2019-summary,sharma-etal-2018-tackling,mostafazadeh-etal-2016-corpus}. 
There is evidence that cloze tasks can be used to effectively pretrain or finetune language models in order to perform few shot learning \citep{schick-schutze-2021-exploiting,liu2021pre}.

WikiHow is a community-edited open domain repository that hosts how-to articles on a variety of different subjects. It is possible to track edits made by users and compare different versions. There have been some recent computational works exploring this resource, including modelling of revision requirements \citep{bhat-etal-2020-towards}, and the effect of edits on fluency \citep{anthonio-roth-2020-learn} and vagueness \citep{debnath-roth-2021-computational}.

\subsection{Masked Language Modelling}
\label{mlm}

Masked Language Modeling (MLM) is a pre-training task which is widely used in transformer-based models. This task forms a self-supervised cloze test by randomly removing some of the tokens of the sentence which will be further filled by the model during training. Prominent transformers including BERT  \citep{devlin-etal-2019-bert} , RoBERTa \citep{Liu2019RoBERTaAR}, DeBERTa \citep{he2020deberta} and T5 \citep{raffel2019exploring} are trained using MLM as an auxiliary objective. Given the successes of transformers in most areas of machine learning and NLP, it is standard practice to fine-tune them for various down-stream tasks.

\subsection{Ordinal Regression}
\label{ordinal}
Ordinal Regression (also known as Ordinal Classification) is a type of classification in which labels have order with respect to each other. Examples of ordinal regression tasks include age estimation \citep{Niu_2016_CVPR}, assessment of damage \citep{ci2019assessment}, and monocular depth estimation \cite{fu2018deep}. In ordinal regression, performance of the model is sensitive to the order of the predictions with regards to the labels. For instance, in the age estimation task, the error of the model should be higher when it incorrectly predicts the age of $30$ as $10$, as opposed to when it predicts the age of $30$ as $20$. Ordinal Regression is commonly done by breaking the multi-class classification task into several binary classification subtasks within a multi-task learning scenario. The output of these binary classification subtasks should be rank-consistent to achieve good performance. 

\section{System Description}
\label{system-desc}

\begin{figure}[t!]
\centering
\includegraphics[width=0.50\textwidth]{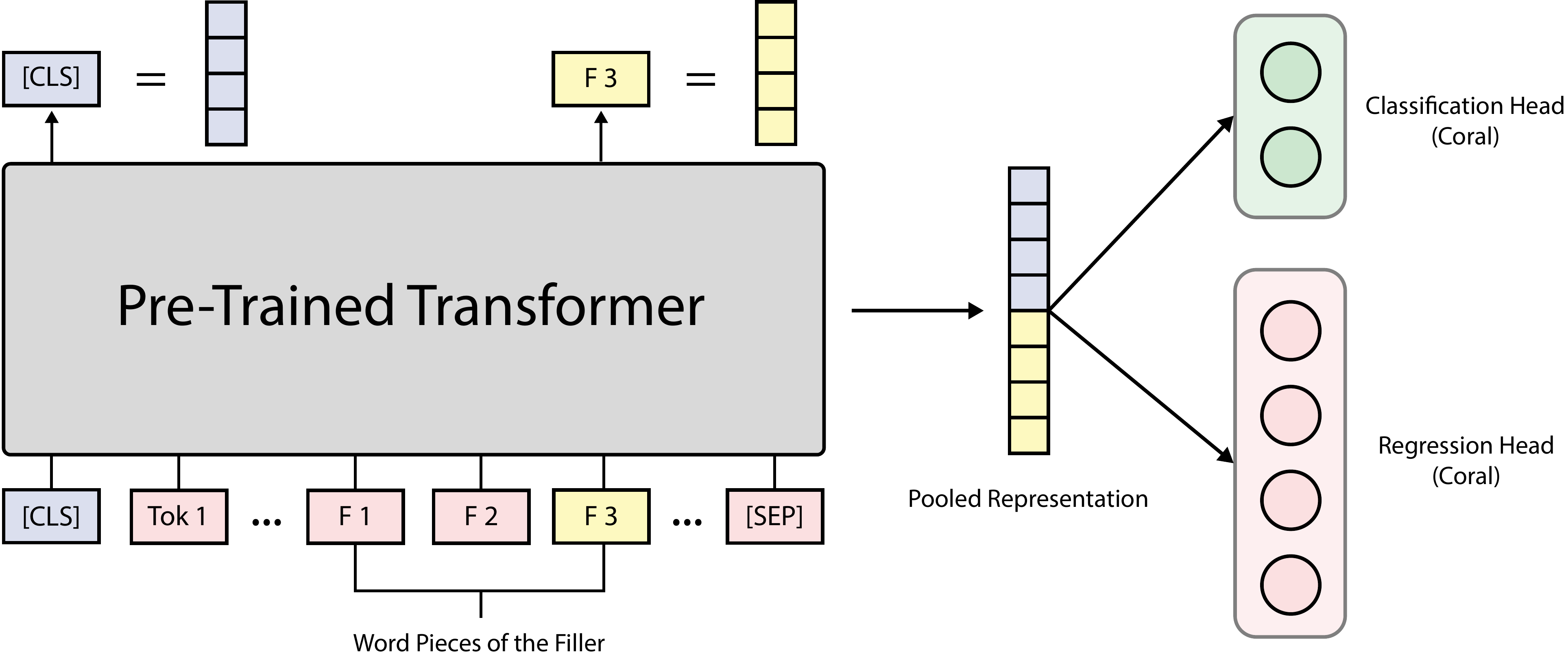}
\caption{Overall architecture of the model. Note how a pooled representation is produced using the last token of the filler and the CLS token}
\label{model-architecture}
\end{figure}

Figure \ref{model-architecture} shows the overall architecture of our model. A pre-trained transformer sits at the base of the architecture. Different variations of this transformer are tried and reported in Sec \ref{results}. These include BERT \citep{devlin-etal-2019-bert}, RoBERTa \citep{Liu2019RoBERTaAR}, DeBERTa (v1 \& 3) \citep{he2020deberta,he2021debertav3}, and T5\footnote{Since T5 is originally an encoder-decoder language model, we only use its pre-trained encoder for our experiments} \cite{raffel2019exploring} in different configurations. 

At the bottom of the network, there are two components, one for regression and the other for classification. At training time, both of these are trained in tandem using ordinal regression and a combined loss. The loss function ensures that the ranks of the labels are kept consistent across these task (Sec. \ref{loss}).

\subsection{Representing the Filler}
To classify each filler, it is first placed in the location of the blank in the text to form a full context. The pre-trained transformer is equipped with a word-piece tokeniser that breaks down the filler to subword units and also uses a [CLS] token to represent the entire context \footnote{T5 does not use CLS for context representation. In that case we just classify the last word-piece with no additional concatenation}. To build a pooled representation for the filler, the contextualised representations for the last word piece and the entire context (i.e. representation of [CLS]) are concatenated and passed on to the next layer. 

\subsection{Multi-task learning with Ordinal Regression}
Once a combined contextualised representation is obtained from concatenation of [CLS] and the filler, we address both subtasks of the shared task in a single multi-task learning architecture. The objective of this model is to predict both the class and the suitability score for each provided filler. The concatenated representation is fed to a fully connected feed-forward layer followed by a GELU activation function \citep{DBLP:journals/corr/HendrycksG16}. This layer projects the representation to a lower dimensional space. Subsequently, the output is passed on to two separate classification and regression heads. 

\subsection{Decomposing the Problem into Binary Classification tasks}
\label{decompose}
As mentioned in Sec. \ref{system-desc}, our model uses a special loss function named coral to perform ordinal regression. This loss function and the training procedure it requires are explained in \citet{cao2020rank}. Coral layers break a $\mathrm{K}$-class classification problem into $\mathrm{K}-1$ binary classification tasks as part of a multi-task learning scenario. For a coral layer with $\mathrm{K}-1$ units \footnote{A $\mathrm{T}$-unit coral layer, is comprised of $\mathrm{T}$ binary classification units which share the same weights but have different biases.}, the value of loss is constructed from the sum of $\mathrm{K}-1$ separate binary cross entropy losses belonging to each unit.

To train a coral layer, it is necessary to first transform the original labels to sets of binary labels. This is the step where the notion of order is introduced into the model. Given $\mathrm{K}$ classes, we convert each label to a collection of binary labels as follows:
\begin{align}\label{eq:labels}
f(y , k)= \begin{cases}
          1 \quad &\text{if}\space\, k < y \\
          0 \quad &\text{if}\space\, k \ge y \\
          \end{cases}
\end{align}
\begin{align*}
Y^{(i)}_{ordinal} = \{ f(y^{(i)} , k)\space|\space k \in \mathbb{W} \space\land\space k < \mathrm{K}-1 \}
\end{align*}

where $y^{(i)}$ is the original label for the $i$th training instance and $0\leq  y^{(i)}\leq \mathrm{K}-1 $.   

\subsection{Classification and Regression Heads}
\label{heads}

At the top end of the architecture there are two components for classification and regression. The classification head is a coral layer with two units which is used to address the classification subtask where we have three labels, namely, \texttt{Implausible}, \texttt{Neutral}, and \texttt{Plausible}. Note that the labels in the classification task have inherent order and for the model to be trained effectively, it is important that the training objective penalises misclassifications based on this underlying assumption. For instance, the error of the model should be more when it predicts an \texttt{Implausible} sample as \texttt{Plausible} compared to when it predicts an \texttt{Implausible} sample as \texttt{Neutral}. This is the motivation to use ordinal regression losses like coral. 

The other head is assigned to the regression subtask and rates each filler for suitability. We converted continuous scores in the $1-5$ range to labels with discrete values in the $0-4$ range by either rounding or flooring the scores. For example, the original scores of $\{1.333, 1.75, 2.5, 3.75, 4.25\}$ are mapped to $\{0, 1, 2, 3, 3\}$ by rounding and $\{0 , 0 , 1 , 2 , 3\}$ by flooring. We can frame the regression task as ordinal classification with five labels by binning the values. A $4$-unit coral layer is used to perform this classification. These heads are jointly trained using a combined loss. 

\subsection{Constructing Labels and Ranks from Heads}
Since coral layers differ from normal dense layers, their output can not be directly converted to labels. Furthermore, the regression subtask is also framed as an additional classification task in our methodology. However, since the purpose is to report continuous scores rather than discrete labels, a unique conversion is necessary for the output of the regression head.

For the classification head, the goal is to output discrete labels in range of $0-2$. Given the output of the classification head for one sample $\hat{C}$, the conversion to the label is defined as follows:
\begin{align}
    f(y)= \begin{cases}
          1 \quad &\text{if}\space\, y > 0.5\\
          0 \quad &\text{if}\space\, y \le 0.5
          \end{cases}
\end{align}
\begin{align}
    c = \sum_{k=1}^{2}f(\sigma(\hat{C}_k))\nonumber
\end{align}

where $c$ is the final label and $\sigma$ is the sigmoid function.\\\\
For the regression head, the goal is to output continuous scores in range of $1-5$. Given the output of the regression head for one sample $\hat{R}$, the conversion to the continuous scores in range of $1-5$ is defined as follows:
\begin{align}
r = (\sum_{k=1}^{4}\sigma(\hat{R}_k)) + 1
\end{align}

\subsection{Computation of Loss}
\label{loss}

Since we have three labels for the classification task, ordinal labels would be sets of binary values with the length of two. Given $C^{(i)}$ as the converted classification label for the $i_{th}$ training sample and the $\hat{C}^{(i)}$ as the output of the classification head for the $i_{th}$ training sample, we can define the classification loss as:
\begin{align}
l^{(i)}_{c} = \sum_{k=1}^{2}-C^{(i)}_{k}\log(\hat{C}^{(i)}_{k})\\\nonumber
-(1-C^{(i)}_{k})\log(1-\hat{C}^{(i)}_{k})\label{eq:loss}
\end{align}

For the regression task there are $5$ different labels, and accordingly, ordinal labels would be sets of binary values with the length of four. Given $R^{(i)}$ as the converted regression label for the $i_{th}$ training sample and the $\hat{R}^{(i)}$ as the output of the regression head for the $i_{th}$ training sample we can define the regression loss in a similar way:
\begin{align}
l^{(i)}_{r} = \sum_{k=1}^{4}-R^{(i)}_{k}\log(\hat{R}^{(i)}_{k})\\\nonumber 
-(1-R^{(i)}_{k})\log(1-\hat{R}^{(i)}_{k})\label{eq:combinedloss}
\end{align}

Since we are aiming to perform both classification and regression, a joint loss is needed to combine losses from the two heads. For a given training batch of size $n$:
\begin{align}
L_{total} = \frac{1}{n}\sum_{i=1}^{n}\lambda_{c}l^{(i)}_{c}\space+\space\lambda_{r}l^{(i)}_r
\end{align}
where $\lambda_c$ is the weight associated with the classification loss and the $\lambda_r$ is associated with regression.

\section{Dataset and Experimental Setup}
\label{experiments}

The shared task is based on \citet{anthonio-etal-2020-wikihowtoimprove}, where instructional texts from the website Wikihow \footnote{A wiki-style online collection of how-to articles accessible at \url{https://www.wikihow.com}} are used to create training instances for a cloze task. The deletions in the dataset are based on actual edits made by online users and they are assumed to represent certain types of clarifications to make a point more clear or disambiguate a sentence. Based on the actual edits, alternative fillers are automatically extracted and added to build a number of possible fillers. These texts along with the fillers were later annotated by humans and given plausibility scores from $1$ to $5$. There is also a separate type of annotation available in which there are $3$ labels with discrete plausibility values. The shared task was organised in two separate tracks of classification and regression, depending on what kind of annotation was used for modelling the task. Table \ref{t:data_subtaskA} \& \ref{t:data_subtaskB} show the basic statistics of the dataset in the shared task.

The dataset for this task consists of six features, named Resolved Pattern, Article title, Section header, Previous context, Sentence, and Follow-up context along with five different fillers for each sample. Resolved Pattern is one of the four following categories: \texttt{IMPLICIT REFERENCE}, \texttt{ADDED COMPOUND}, \texttt{METONYMIC REFERENCE}, and \texttt{FUSED HEAD} which indicates the relationship of the fillers with the context. Article title is the name of the article from which the paragraph is selected. Section header is the section from which the article is selected. Previous context is a few sentences before the sentence that contains the filler. Sentence is the sentence that contains the filler.  

We obtained our best results when we used a custom formatting using which we can feed all the features to the model as textual input. Table \ref{t:input_formatting} is an example of how we represent each training instance. Note how the `Text' feature is constructed by concatenating previous and follow-up contexts with the target sentence. 

\begin{table}[ht!]
    \smallskip
    \small
    \centering
    \begin{tabular}{p{\linewidth-10pt}}
        \toprule[1pt]
        \textbf{Example input while using all features}
        \\\midrule[0.5pt]
        Resolved pattern: ADDED COMPOUND\\
        Section header: Following a Basic Routine\\
        Article title: How to Get Rid of Peeling Skin\\
        Text: (...) 6. Never tear away loose skin. (...) 7. Protect your skin from sunlight. Exposure to direct sunlight can weaken your skin further and complicate the [Filler] problem. This is true regardless of whether your skin is peeling due to a sunburn or due to dryness.
        \\\bottomrule[0.5pt]
    \end{tabular}
    \caption{\label{t:input_formatting} An example of how each input is formatted for training and inference. Each identifier followed by ":" represents a feature of the dataset represented in textual format.}
\end{table}

\begin{table}[ht!]
\smallskip
\small
\begin{tabular}{c|cccc}
\toprule
 Variation & implausible & neutral & plausible & total\\
\midrule
train   & 5474 & 7162 & 7339 & 19975\\
        & (27.40\%) & (35.85\%) & (36.75\%) & (100\%)\\
        \cmidrule{1-5}
dev     & 982 & 602 & 916 & 2500\\
        & (39.28\%) & (24.08\%) & (36.64\%) & (100\%)\\
\hline
\end{tabular}
\caption{\label{t:data_subtaskA} Statistics of the data for subtask A}
\end{table}

\begin{table}[ht!]
\smallskip
\small
\begin{tabular}{l|cccccc}
\toprule
Variation & 1 & 2 & 3 & 4 & 5 & total\\
\midrule
train\hspace{0.25cm}F & 2254 & 4123 & 6259 & 5321 & 2018 & 19975\\
train\hspace{0.25cm}R & 1053 & 4421 & 4034 & 8441 & 2026 & 19975\\
dev\hspace{0.3cm}  F & 645  & 458  & 481  & 596  & 320 & 2500\\
dev\hspace{0.3cm}  R & 386  & 596  & 376  & 639  & 503 & 2500\\
\hline
\end{tabular}
\caption{\label{t:data_subtaskB} Statistics of the data for subtask B. R and F represent rounded and floored scores, respectively}
\end{table} 

\section{Results and Discussion}
\label{results}

The official test results for both subtasks are presented in Table \ref{t:results_leaderboard}. These are the best results that we have obtained on the test set prior to the end of the competition. On the official leaderboard of the shared task, our system was ranked $5$th and $7$th for the ranking and classification tasks, respectively. 

We have since performed extensive analyses on the effect of hyperparameters on different variations of our models, and since we do not have access to the true labels in the test set, we report our best results on the dev set. As can be seen in Table \ref{t:results}, we have produced our best results using DeBERTa-V3$_{large}$. We have noticed two important factors that influence the final performance of the models. The first factor is batch size. Our  best results were obtained on lower batch sizes of $8$ and $16$. The choice of rounded or floored numbers for subtask B is also a significant factor. The reason for this is that the distribution of labels changes depending on the normalisation method used. 

\begin{table*}[ht!]
    \centering
    \begin{tabular}{P{2.5cm}|p{3cm}P{1.5cm}P{1.5cm}|P{2cm}|P{3cm}}
        \toprule[1pt]
        Backbone Transformer & Model & Batch Size & Scores  & Accuracy (subtask A) & Spearman's rank (subtask B) \\\midrule[1pt]
                   & Bert$_{base}$ & 8 & R & 57.80\% & 0.6020\\
                   & Bert$_{base}$ & 8 & F & 57.48\% & 0.6038\\
                   & Bert$_{base}$ & 16 & R & 56.68\% & 0.5950\\
        BERT       & Bert$_{base}$ & 16 & F & 57.92\% & 0.5960\\
                   & Bert$_{large}$ & 8 & R & 57.60\% & 0.6323\\
                   & Bert$_{large}$ & 8 & F & 59.12\% & 0.6341\\
                %   & Bert$_{large}$ & 16 & R & - & -\\
                %   & Bert$_{large}$ & 16 & F & - & -\\
                   \cmidrule[0.8pt]{1-6}
                   & RoBERTa$_{base}$ & 8 & R & 59.16\% & 0.6610\\
                   & RoBERTa$_{base}$ & 8 & F & 58.68\% & 0.6582\\
                   & RoBERTa$_{base}$ & 16 & R & 57.32\% & 0.6743\\
        RoBERTa    & RoBERTa$_{base}$ & 16 & F & 59.44\% & 0.6695\\
                   & RoBERTa$_{large}$ & 8 & R & 58.68\% & 0.6799\\
                   & RoBERTa$_{large}$ & 8 & F & 60.20\% & 0.6928\\
                %   & RoBERTa$_{large}$ & 16 & R & - & -\\
                %   & RoBERTa$_{large}$ & 16 & F & - & -\\
                   \cmidrule[0.8pt]{1-6}
                   & DeBERTa-V3$_{base}$ & 8 & R & 61.72\% & 0.7194\\
                   & DeBERTa-V3$_{base}$ & 8 & F & 62.64\% & 0.7260\\
                   & DeBERTa-V3$_{base}$ & 16 & R & 63.12\% & 0.7287\\
        DeBERTa-V3 & DeBERTa-V3$_{base}$ & 16 & F & 63.84\% & 0.7326\\
                   & DeBERTa-V3$_{large}$ & 8 & R & \textbf{64.12\%} & 0.7406\\
                   & DeBERTa-V3$_{large}$ & 8 & F & \textbf{64.12\%} & \textbf{0.7411}\\
                %   & DeBERTa-V3$_{large}$ & 16 & R & - & -\\
                %   & DeBERTa-V3$_{large}$ & 16 & F & - & -\\
                   \cmidrule[0.8pt]{1-6}
                   & T5-Encoder$_{base}$ & 8 & R & 58.88\% & 0.6598\\
                   & T5-Encoder$_{base}$ & 8 & F & 59.48\% & 0.6573\\
                   & T5-Encoder$_{base}$ & 16 & R & 59.24\% & 0.6526\\
        T5         & T5-Encoder$_{base}$ & 16 & F & 59.08\% & 0.6568\\
                   & T5$_{large}$ & 8 & R & 61.76\% & 0.6949\\
                   & T5$_{large}$ & 8 & F & 62.24\% & 0.6907\\
                %   & T5$_{large}$ & 16 & R & - & -\\
                %   & T5$_{large}$ & 16 & F & - & -\\
                   \bottomrule
    \end{tabular}
    \caption{\label{t:results} Results on dev set for both subtasks, R and F represent rounded and floored scores respectively (as mentioned in Sec. \ref{heads})}
\end{table*}

\begin{table*}[ht!]
    \centering
    \begin{tabular}{p{3.5cm}P{2.5cm}P{2.5cm}|P{2.5cm}|P{3cm}}
        \toprule[1pt]
        Model & Batch Size & Scores  & Accuracy (subtask A) & Spearman's rank (subtask B) \\\midrule[1pt]
                   RoBERTa$_{large}$ & 8 & F & 61.00\% & 0.6700\\
                   DeBERTa-V1$_{large}$ & 8 & F & 61.00\% & 0.6900\\
                   T5-Encoder$_{large}$ & 16 & F & \textbf{62.40}\% & \textbf{0.7070}\\
        \bottomrule
    \end{tabular}
    \caption{\label{t:results_leaderboard} Official results on leaderboard for both subtasks}
\end{table*}

% \section{Error Analysis}
% \label{error-analysis}

\section{Conclusion}
\label{Conclusion}

In this work we developed a set of transformer-based models powered with ordinal regression to tackle an NLP cloze task as part of the SemEval 2022 shared task 7. The goal was to assign suitability scores or labels to several different provided fillers given each context and masked span. 

Using a combined architecture based on ordinal regression that used training labels from both subtasks, we developed and trained models with a multi-task learning objective. The proposed system was ranked $5$th out of 21 teams in the ranking track of the shared task. In the subsequent analyses in the post evaluation phase, we have showed the effectiveness of this architecture in addressing this task. We compared different variations of our models and explored the effects of hyperparameters on model performance. The code and analyses are going to be publicly available.   

\bibliography{anthology,custom}

\appendix
\section{Appendix}
In this section, we are going to share the details of the hyperparameters used for the fine-tuning of our models and the final training procedure used for the submission.

\subsection{Hyperparameters}
We fine-tuned all models for $5$ epochs while keeping the embedding layers of all models frozen. We used the AdamW as our optimizer with a cosine learning rate schedular from the Hugging Face library and a weight decay of $0.00123974$ and an initial learning rate of $1.90323e-05$. Also, the $\lambda_l$ and $\lambda_r$ (as mentioned in \ref{loss}) are set to $0.5$ in all of the reported experiments as shown in Table \ref{t:hyperparams}.

\begin{table}[h!]
    \begin{tabular}{p{(\linewidth-20pt)-65pt}P{60pt}}
        \toprule[1pt]
        \textbf{Hyperparamters}&
        \\\midrule[0.5pt]
        Epochs&5\\
        Optimizer&AdamW\\
        Learaning Rate Scheduler&Cosine\\
        Initial Learning Rate&1.90323e-05\\
        Weight Decay&0.00123974\\
        $\lambda_l$&0.5\\
        $\lambda_r$&0.5
        \\\bottomrule[0.5pt]
    \end{tabular}
    \caption{\label{t:hyperparams}Details of the Hyperparamters used for fine-tuning}
\end{table}

\subsection{Training Procedure for Submission}
Once we found the best hyperparameters and models using Dev dataset, We used a combination of Training and Dev data to train our final models for submission. With this approach, we have been able to achieve a $3\%$ improvement on accuracy and up to $0.05$ improvement on Spearman’s rank correlation. Additionally, we found that combining Train and Dev data is noticeably less effective in the smaller models such as Bert$_{base}$ or RoBERTa$_{base}$, however, this is a better strategy when it comes to larger models such as RoBERTa$_{large}$ or T5$_{large}$. Based on this, we suppose that with further tuning. DeBERTa-V3$_{large}$ can potentially surpass the state-of-the-art if it has access to the combination of Train and Dev data.
\end{document}